\documentclass{llncs}
\usepackage{graphicx}
\usepackage{xcolor}
\usepackage{adjustbox}
\usepackage{colortbl}
\usepackage{array}

\usepackage{arydshln}
\usepackage{tikz}
\usetikzlibrary{shapes,arrows}
\usepackage{xspace}
\usepackage{todonotes}
\newcolumntype{P}[1]{>{\centering\arraybackslash}p{#1}}

\newcommand\Secref[1]{\mbox{Sec.}~\ref{#1}}
\newcommand{\Tabref}[1]{\mbox{Tab.}~\ref{#1}}
\newcommand\Figref[1]{\mbox{Fig.}~\ref{#1}}
\newcommand\etal{\mbox{\textit{et al.}}\xspace}

\title{Closing the Reality Gap with Unsupervised Sim-to-Real Image Translation}

\author{Jan Blumenkamp\inst{1}, Andreas Baude\inst{2} and Tim Laue \inst{2}}

\institute{University of Cambridge, Department of Computer Science and Technology, \\
Cambridge, United Kingdom \\
\email{jb2270@cam.ac.uk} \and
Universität Bremen, Fachbereich 3 -- Mathematik und Informatik, \\Postfach 330 440, 28334 Bremen, Germany \\
\email{$\{$an\_ba,tlaue$\}$@uni-bremen.de}}

\begin{document}

\maketitle

\begin{abstract}
    Deep learning approaches have become the standard solution to many problems in computer vision and robotics, but obtaining sufficient training data in high enough quality is challenging, as human labor is error prone, time consuming, and expensive.
    Solutions based on simulation have become more popular in recent years, but the gap between simulation and reality is still a major issue.
    In this paper, we introduce a novel method for augmenting synthetic image data through unsupervised image-to-image translation by applying the style of real world images to simulated images with open source frameworks.
    The generated dataset is combined with conventional augmentation methods and is then applied to a neural network model running in real-time on autonomous soccer robots.
    Our evaluation shows a significant improvement compared to models trained on images generated entirely in simulation.
    \end{abstract}

    \section{Introduction}

In recent years, deep learning approaches became the standard solution to many problems in computer vision, such as classification \cite{ciresan_2010}, object detection \cite{ren_2015}, or semantic segmentation \cite{ronneberger_2015}. Efforts were made to reduce the computational complexity in order to deploy them to mobile devices \cite{howard_2017}.
These approaches usually require a vast amount of training data which can either be generated through human labor or be generated synthetically.
Generating training data through human labor can result in datasets of high quality, but it is a cumbersome and expensive task.
The CamVid dataset \cite{brostow_2009} contains detailed semantic labels and uses preprocessing to assist human labelers, but annotating a single frame takes 20 to 25 minutes.
Multiple volunteers were tasked to perform the labeling, but only about 15 \% of the volunteers delivered acceptable results.
Similar problems exist in other datasets such as the PASCAL VOC challenge \cite{everingham_2010} or in the COCO dataset \cite{lin_2014}.

Recently, a trend can be seen to approaches that rely on simulated and synthetic data.
The SYNTHIA dataset, for instance, consists of $213400$ images and pixel-accurate semantic annotations as well as depth maps generated with the Unity framework \cite{ros_2016}.
Computer games can also be used to generate images that can then be labeled manually \cite{richter_2016}.
Unfortunately, data generated in a simulated environment often does not directly transfer to reality.
This issue is referred to as the \textit{reality gap} \cite{jakobi_1995}.

Hess \etal \cite{hess_2018} introduced an environment to create annotated training data in a RoboCup Standard Platform League (SPL) setting and demonstrated the feasibility of performing a semantic segmentation on that data.
A major challenge in the RoboCup SPL is the perception of the field in a diverse set of lighting conditions.
Low quality cameras result in images with low contrasts and limited processing power usually requires using fast conventional computer vision approaches. Frameworks such as TensorFlow Lite \cite{abadi_2015} or CompiledNN \cite{thielke_2019} made utilizing neural networks in mobile and low-end devices more feasible.

In our work, we synthetically generate images with the tools provided by \cite{hess_2018} and transform them with unsupervised image-to-image translation \cite{huang_2018} and domain randomization \cite{tremblay_2017} so that they can be used as training data for any kind of deep learning task.
We use this dataset to train a semantic segmentation that is able to run in real-time on a NAO v6 robot.
Our approach can generally be applied to any other domain where computing power is sparse and flexibility and reliability plays an important role. All required software dependencies are open source.
Our evaluation shows that models trained with our method perform noticeably better than models trained with data directly generated from simulation as well as with generated data that is expanded with conventional augmentation techniques.
In summary, our main contributions are:
\begin{itemize}
    \item We developed a method using publicly available state of the art image-to-image transformation frameworks and demonstrate that it can be used to generate high quality datasets that allow training highly performing models.
    \item We introduce a multi-class semantic segmentation model architecture that is capable of running in real-time on a NAO v6 robot.
\end{itemize}

\begin{figure}[t]
  \centering
  \definecolor{babyblueeyes}{rgb}{0.63, 0.79, 0.95}
  \tikzstyle{diamo} = [diamond, draw, fill=babyblueeyes!20, text width=1em, text badly centered, inner sep=0pt]
  \tikzstyle{block} = [rectangle, draw, fill=babyblueeyes!20, text width=6em, text centered, rounded corners]
  \tikzstyle{line} = [draw, -latex']
  
  \begin{tikzpicture}[scale=0.8, every node/.style={scale=0.8}]
      \node [block] (uerobocup) {Unreal Engine UERoboCup};
      \node [block, below of=uerobocup, node distance = 35mm] (imAnMas) {Raw images \vspace{0.2em} and masks \vspace{0.1em} \includegraphics[width=\linewidth]{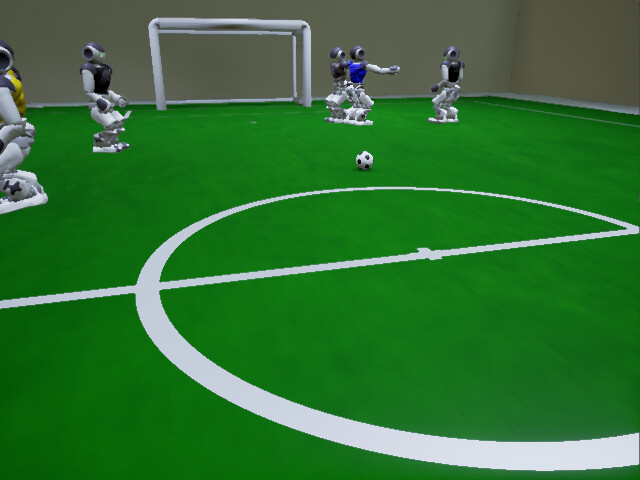} \includegraphics[width=\linewidth]{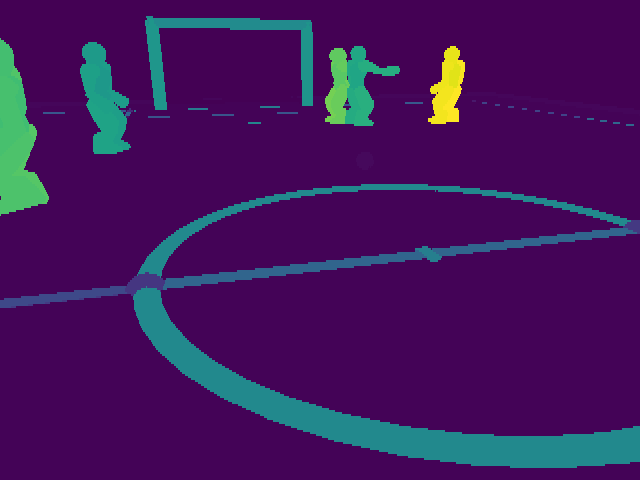}};
       \node [diamo, right of=imAnMas, node distance = 27.5mm] (dec1) {};
      \node [block, below of=dec1, node distance = 22mm] (coco) {\vspace{0.2em} COCO \includegraphics[width=\linewidth]{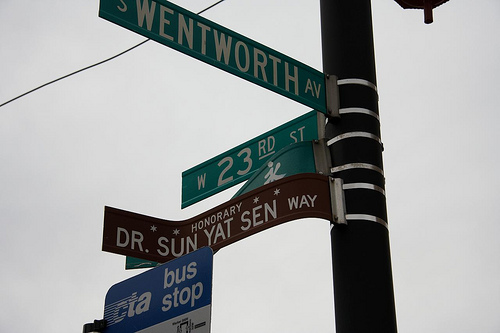}};
      \node [diamo, right of=dec1, node distance = 17mm] (dec2) {};
      \node [block, above of=dec2, node distance = 13mm] (munit) {MUNIT};
      \node [block, above left of=munit, node distance = 26mm] (real) {\vspace{0.2em} Real \includegraphics[width=\linewidth]{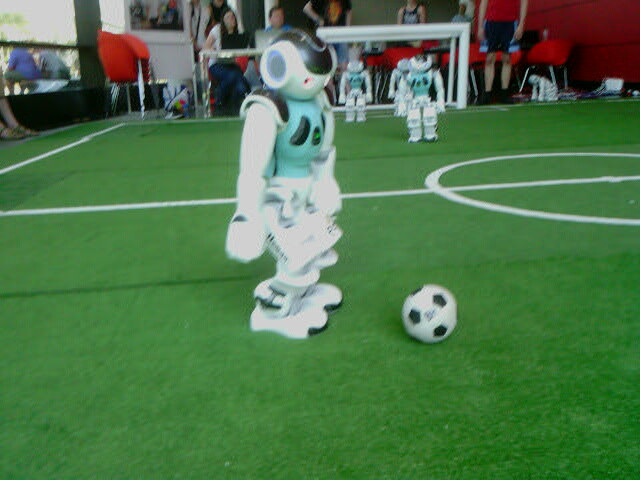}};
      \node [block, above right of=munit, node distance = 26mm] (sim) {\vspace{0.2em} Sim \includegraphics[width=\linewidth]{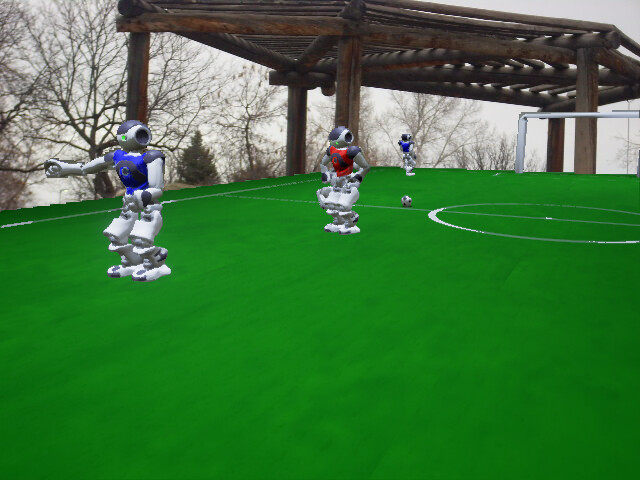}};
      \node [block, right of=dec2, node distance = 22mm] (dataset) {\vspace{0.2em} Generated Dataset};
      \node [block, right of=dataset, node distance = 28mm] (semseg) {Semantic Segmentation};
      \node [block, above of=semseg, node distance = 25mm, text width=6em] (augment) {Augmentation \includegraphics[width=\linewidth]{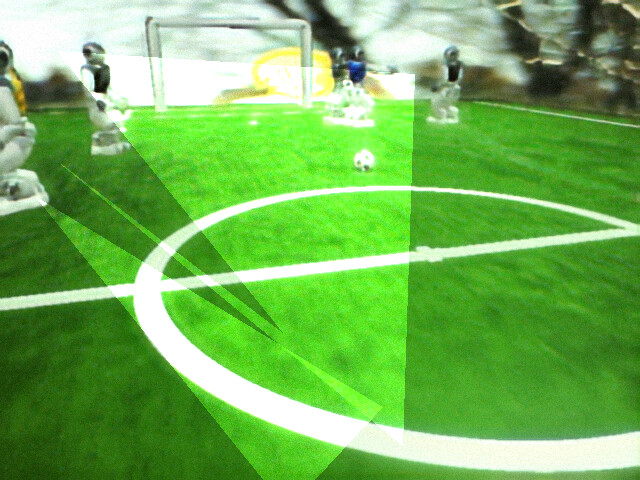}};
      \node [block, right of=semseg, node distance = 28mm] (inference) {NAO};
      \node [block, above of=inference, node distance = 25mm] (input) {Input \includegraphics[width=\linewidth]{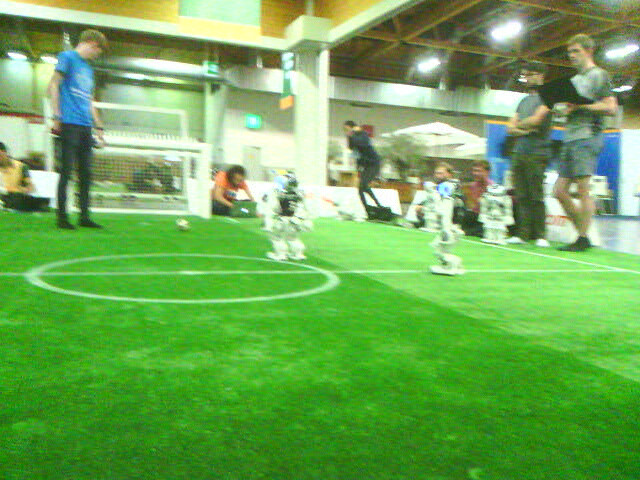}};
      \node [block, below of=inference, node distance = 22mm] (prediction) {\vspace{0.2em} Prediction \includegraphics[width=\linewidth]{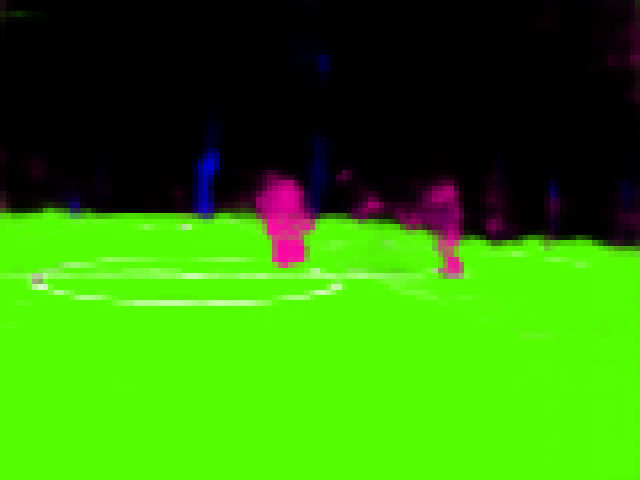}};
  
      \path [line] (uerobocup) -- node [text width=2.7cm,midway] {Generate} (imAnMas);
      \path [line] (imAnMas) -- (dec1);
      \path [line] (coco) -- node [text width=2.2cm,midway, right] {Replace background} (dec1);
      \path [line] (dec1) -- (dec2);
      \path [line] (munit) -- node [text width=1.3cm, midway, left] {Transfor\-mation} (dec2);
      \path [line] (real) -- node [midway, left] {$\mathcal{X}_\mathrm{real}$} (munit);
      \path [line] (sim) -- node [midway, right] {$\mathcal{X}_\mathrm{sim}$} (munit);
      \path [line] (dec2) -- (dataset);
      \path [line] (dataset) -- (semseg);
      \path [line] (augment) -- node [midway, left] {Training} (semseg);
      \path [line] (semseg) -- (inference);
      \path [line] (input) -- (inference);
      \path [line] (inference) -- node [midway, left] {Inference} (prediction);
  \end{tikzpicture}
  \caption{The general workflow of our proposed method: Images and corresponding masks are generated in simulation.
  The background of the generated images is replaced with a random image from the COCO dataset.
  MUNIT image-to-image translation is applied to these enhanced images, which are now the intermediate dataset.}
  \label{fig:workflow}
  \end{figure}
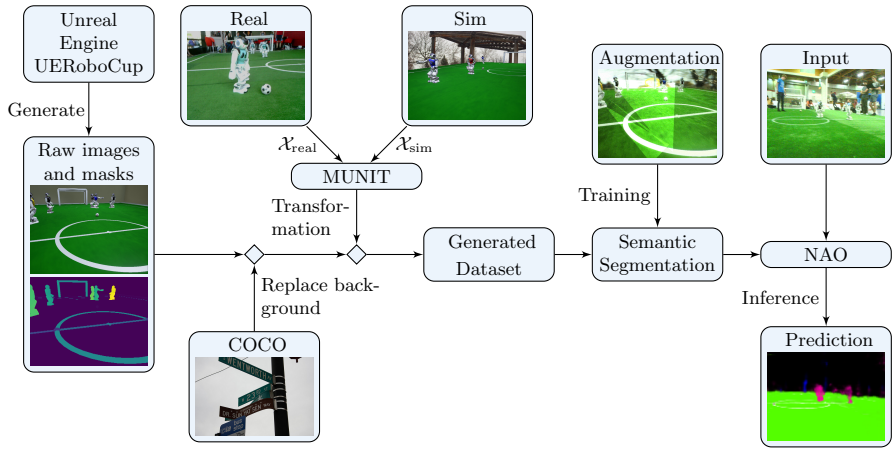

The remainder of this paper is organized as follows: After summarizing the related work in \Secref{s:related_work}, we describe our data generation approach in \Secref{s:data_generation}. In \Secref{s:semantic_segmentation}, we present our semantic segmentation model architecture. Lastly, multiple models generated with different data augmentations are evaluated and discussed in \Secref{s:results} and \Secref{s:discussion} respectively. An overview of the proposed method is depicted in \Figref{fig:workflow}.

\section{Related Work}
\label{s:related_work}

There are two remarkable works in the area of \textit{closing the reality gap with image-to-image translation}:
Bousmalis \etal \cite{bousmalis_2017} use domain adaptation and domain-adversarial neural networks to utilize synthetic training data in an end-to-end learning approach in order to learn robotic grasping.
Bewley \etal \cite{bewley_2018} use image-to-image translation to transfer a vision-based driving policy from simulation to reality.
Instead of explicit representations such as a semantic segmentation, their end-to-end approach uses more implicit representations.
In contrast to our proposed method, both approaches have in common that they utilize the image-to-image translation in an end-to-end approach. Flexibility in the postprocessing plays an important role in RoboCup settings, which is not given with the current state of the art of end-to-end approaches. Using intermediate representations such as pixel-accurate labels allows a higher flexibility and versatility. A dataset using such low level representations can be used both for semantic segmentation and for high level classification and object detection.

\textit{Deep learning approaches for robot vision in RoboCup} were first applied in the humanoid league, where
Schnekenburger \etal \cite{schnekenburger_2017} use a segmentation to detect different field features such as line intersections and objects.
All classes except for the detection of other robots performed satisfactory.
Van Dijk \etal \cite{dijk_2018} propose a novel model architecture without any residual connections for a semantic segmentation, which was able to be executed in close to real-time on a typical smartphone CPU, but lacks the ability to properly detect complex features or multiple classes at once.
In contrast to the humanoid league, the SPL, which uses the SoftBank NAO robots, is more constrained in terms of hardware.
Nevertheless, Hess \etal \cite{hess_2018} trained a simple classifier model to demonstrate the feasibility of using synthetic images in such a scenario.
Szemenyei \etal \cite{szemenyei_2018} propose a novel, small semantic segmentation model that uses images generated with {\em UERoboCup} for pre-training the model and eventually tune it with real images.
However, this approach does not focus on the more extreme conditions that games in environments with natural lighting require.
Furthermore, Poppinga \etal \cite{poppinga_2019} proposed a robot detection framework for which data obtained in a simulation was used to learn additional features that are hard to label manually, such as robot distances.

\section{Data Generation}
\label{s:data_generation}
In this section, after briefly describing the background and the tools generating the simulated data, we give insights into the Multimodal Unsupervised Image-to-Image Translation (MUNIT) framework \cite{huang_2018}, which we use for the sim-to-real image translation, and describe the online data augmentation methods we used.

\subsection{RoboCup and UERoboCup}
While RoboCup provides a benchmark to quantitatively compare the progress over time, the community is small and lacks labeled training data for deep learning approaches.
Efforts were made to create a community-driven database for labeled real images \cite{fiedler_2018}, but the data required for specific tasks and pixel-accurate labels are rarely available.

UERoboCup is an application based on \textit{Unreal Engine} that allows generating game situations with multiple robots from the view of a specific robot \cite{hess_2018} and is capable of generating pixel-accurate semantic annotation.
To provide a more accurate representation of the environment, we added additional labels relevant to the SPL context, such as the penalty mark and the goal bar as well as an adapted appearance of the robots to that of the latest NAO robot generation.
In addition, we increased the variation in the generated images to reflect reality more properly.
This is achieved by a variable camera pitch instead of a fixed one and the definition of a skeleton for the previously static robot mesh, which allows dynamic robot poses.
This procedure is referred to as domain randomization \cite{tremblay_2017}.
Furthermore, the robot skeleton can be used for a more detailed segmentation of individual robot limbs, allowing, for instance, a pose detection.
We also export meta data such as the extrinsic camera parameters and the poses of robots in the standardized JSON format.
Such information is difficult to annotate manually, therefore learning further characteristics, such as the distance to another robot, as shown in \cite{poppinga_2019}, mainly relies on synthetic data.

We generated a set of $10000$ images and labels with UERoboCup.
A generated image with the corresponding converted segmentation mask can be seen in the overview in \Figref{fig:workflow}.

\subsection{Image Post Processing}
UERoboCup only creates images of a plain RoboCup scene taking place in a white room.
However, during an actual game, the background is cluttered with a wide range of different objects, such as people walking around.
In order to make potential deep learning applications understand the concept of unwanted background clutter, we replace the background with structured images, similar to \cite{tremblay_2017}.
This can be considered another variation of domain randomization.
We use images from the COCO test set \cite{lin_2014}.
Even though these images do not exactly represent how a scene would look like at a RoboCup event, we found that they are well-suited to help deep learning applications differentiate between relevant foreground and irrelevant background clutter.

\subsection{MUNIT}
MUNIT assumes images from two different domains $x_1 \in \mathcal{X}_1$ and $x_2 \in \mathcal{X}_2$ and, given samples drawn from the two marginal distributions $p(x_1)$ and $p(x_2)$, without access to the joint distribution $p(x_1, x_2)$, estimate the conditionals $p(x_1|x_2)$ and $p(x_2|x_1)$ with the image-to-image translation models $p(x_{1\rightarrow 2}|x_2)$, where $x_{1\rightarrow 2}$ is a sample resulting from translating $x_1$ to $\mathcal{X}_2$ \cite{huang_2018}.
Due to the unsupervised nature of MUNIT, no explicit labeling has to be performed on any of the sample images.
The learned mapping is multimodal, thus multiple different images with different styles from domain $x_1 \in \mathcal{X}_1$ can be applied to the same image $x_2 \in \mathcal{X}_2$ and each time a different image with the style from domain $x_1 \in \mathcal{X}_1$ is computed.

\subsection{Style Transfer}
\begin{figure}[t]
  \centering
  \begin{minipage}[t]{0.32\textwidth}
    \centering
    \includegraphics[width=\textwidth]{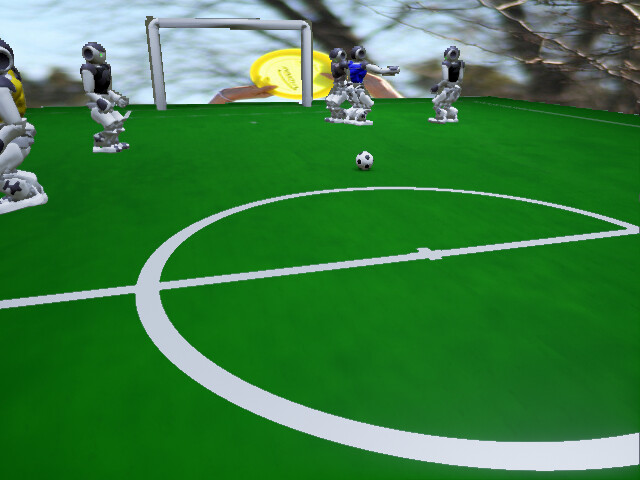}
  \end{minipage}
  \hfil
  \begin{minipage}[t]{0.32\textwidth}
    \centering
    \includegraphics[width=\textwidth]{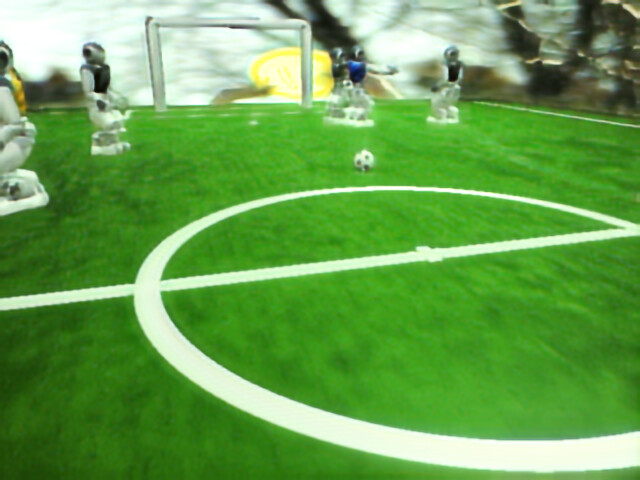}
  \end{minipage}
  \hfil
  \begin{minipage}[t]{0.32\textwidth}
    \centering
    \includegraphics[width=\textwidth]{img/pipeline/6_online_aug.jpg}
  \end{minipage}
  \caption{After changing the background in the simulated image to a random image from the COCO dataset (left), we apply the image-to-image translation learned by MUNIT to it (middle).
In addition to the offline augmentation obtained with the sim-to-real translation, we perform an online augmentation during training (right).
Beside standard augmentations such as blur, we also introduce a simple simulation of sun light patches on the field.}
  \label{fig:augmentations}
\end{figure}

MUNIT requires a test set and a training set of images for the two classes $\mathcal{X}_1$ and $\mathcal{X}_2$.
For this application, the two classes are real images recorded by a robot's camera $\mathcal{X}_\mathrm{real}$ and simulated images created with UERoboCup $\mathcal{X}_\mathrm{sim}$.
For the real domain, we select images from previous RoboCup events and from ImageTagger \cite{fiedler_2018}, accumulating to an overall of $885$ images in the training set and $155$ images in the test set.
We found that a large variance in the training images is essential for MUNIT to generate useful results.
Since the images generated by UERoboCup are random, any subset with about the same size can be used.
Note that in this subset, we already replaced the background with a random image from the COCO dataset.
In the simulated domain, we used $1000$ images for training and $200$ for testing.
We train on an NVidia Titan V with MUNIT default settings for $70000$ epochs.
Due to memory limitations, we slightly decreased the amount of the generator and discriminator filters.
We noticed a convergence after $50000$ epochs, with no further significant improvements from that point.
To generate the processed images, we took three random style images from the set of real images and applied them to each image generated by UERoboCup, resulting in $30000$ different images.
An example can be seen in \Figref{fig:augmentations}.

\subsection{Data Augmentation}
\label{sec:augmentation}
Data augmentation is a type of data regularization and helps to avoid overfitting.
Additionally, data augmentation can be used to enhance the size and quality of datasets with warping or oversampling methods \cite{shorten_2019}.

We used the following online data augmentations during training: Vertical image flipping, Gaussian noise, multiply, add (RGB and HSV), simplex noise, motion blur, contrast normalization, and simulated sun patches.

It is essential that the model learns to handle extreme environmental situations with patches of light and shadow.
Since this is not captured with sufficient variance in our dataset, we introduce an additional domain randomization method during online data augmentation by simulating patches on the field that are illuminated by the sun.
We implement this augmentation by generating multiple random polygons consisting of three to six points in the frame and multiply these areas with a random factor.
A sample augmentation is shown in \Figref{fig:augmentations}.

\section{Semantic Segmentation}
\label{s:semantic_segmentation}
Designing models that are capable of being executed in real-time on platforms with limited resources is difficult due to computing constraints.
In our application, the performance baseline for the design of the model is the NAO's camera frame rate of $30$~fps.
We allocate one CPU core for the processing of the frames of one camera, which means that the model must process each frame at the same rate.
As inference framework, we use \textit{CompiledNN} \cite{thielke_2019}.

The model architecture is based on the U-Net architecture \cite{ronneberger_2015} and incorporates features from MobileNet \cite{howard_2017}.
The U-Net architecture consists of an encoder and a decoder with multiple residual connections between the encoder and the decoder.
The reduced model from this work uses only two residual connections and downscales the image twice.
MobileNet was designed to be deployed to mobile devices and is therefore optimized to run with limited computing power.
This is mainly realized by replacing convolutions with separable convolutions \cite{howard_2017}.
For an additional acceleration, pooling layers are replaced with convolutions with a corresponding stride, as proposed in \cite{springenberg_2015}.
We use batch normalization for regularization \cite{ioffe_2015} and LeakyReLU \cite{xu_2015} as activation functions.
The resulting model has $12909$ learnable parameters and is shown in \Tabref{tab:network}.

Our proposed model is capable of performing multi-class predictions (in our case ball, goal posts, lines, and robots).
Due to limitations in CompiledNN, no softmax activation is applied to the last layer. This means that the model detects all mentioned classes independently from each other. Note that a binary cross entropy or something similar must be used as the loss function.
Furthermore, it can be desirable to have multiple independent classifications outputs.
For instance, if the model is not certain if the arm of a robot is a similarly looking goal post, the likelihood of both outputs would be small with a softmax activation.
By keeping the output layers independent from each other, the model can classify the arm of a robot both as part of a robot but also as a goal post.
This can be validated in the postprocessing by applying additional domain knowledge and thus discarding false positives. Note that ideally, the model is able to capture this from context, but we found that increasing the model size more leads to an unacceptably low inference time.
The maximum amount of features is limited by the complexity of the model. We achieved good results with predicting five and less classes, but it is to be expected that a larger amount of independent output predictions yields worse predictions, if the base model is not adapted.

\begin{table}[t]
\centering
\caption{Our proposed segmentation model architecture. The scale refers to the tensor size relative to the input size. N is the amount of how often the layer in the row is repeated and F represents the amount of filters.}
\label{tab:network}
\begin{tabular}{l|l|l|l}
\textbf{Layer} & \textbf{Scale} & \textbf{N} & \textbf{F} \\
\hline
Input & 1 & 1 & 3 \\
$3 \times 3$-SConv2D, BN, LeakyReLU & 1 & 1 & 8 \\
$\phantom{|}\vline$\qquad $3 \times 3$-SConv2D, BN, LeakyReLU & 1  & 1 & 8 \\
$\phantom{|}\vline$\qquad $3 \times 3$-SConv2D, BN, LeakyReLU & 1/2 & 1& 8 \\
$\phantom{|}\vline$\qquad$\phantom{|}\vline$\qquad $3 \times 3$-SConv2D, BN, LeakyReLU & 1/2 & 2 & 16 \\
$\phantom{|}\vline$\qquad$\phantom{|}\vline$\qquad $3 \times 3$-SConv2D, BN, LeakyReLU & 1/4 & 1 & 16 \\
$\phantom{|}\vline$\qquad$\phantom{|}\vline$\qquad $3 \times 3$-SConv2D, BN, LeakyReLU & 1/4 & 6 & 24 \\
$\phantom{|}\vline$\qquad$\phantom{|}\vline$\qquad Up2D & 1/2 & 1& 24 \\
$\phantom{|}\vline$\qquad Concat & 1/2 & 1 & 40 \\
$\phantom{|}\vline$\qquad $3 \times 3$-SConv2D, BN, LeakyReLU & 1/2 & 3 & 16 \\
$\phantom{|}\vline$\qquad Up2D & 1 & 1 & 16 \\
Concat & 1 & 1 & 24 \\
$3 \times 3$-SConv2D, BN, LeakyReLU & 1 & 3 & 8 \\
$3 \times 3$-SConv2D, BN, LeakyReLU & 1 & 1 & 5
\end{tabular}
\end{table}

We perform different experiments to find a working model architecture and look for a compromise between detection accuracy and inference time.
The runtime was measured on a single CPU core of the NAO v6 (Intel Atom E3845@1.91 GHz \cite{softbank_nao_2018}) using CompiledNN \cite{thielke_2019}. The results for different image resolutions are shown in \Tabref{tab:execution_time}.
As the maximum feasible input resolution for real-time operation is 14 ms, the original input image is subsampled accordingly. The resulting aliasing should not affect the neural model as it should learn to ignore it.

\begin{table}[b]
  \centering
  \caption{Inference time on the NAO v6 with CompiledNN}
  \label{tab:execution_time}
  \begin{tabular}{c|c|c|c|c|c|c}
    Size {[}px $\times$ px{]} & $40 \times 32$ & $80 \times 64$ & $108 \times 80$ & $120 \times 88$ & $160 \times 120$ & $320 \times 240$ \\ \hline
    Duration {[}ms{]}  & $1.6$     & $6.7$     & $11.2$     & $14.0$       & $27.3$      & $116.0$      
  \end{tabular}
\end{table}

\section{Results}
\label{s:results}
In order to evaluate the performance of the model and the synthetic dataset, we manually labeled 348 images, which we also used in the MUNIT augmentation resembling varying environment conditions with the five classes field, lines, robots, goal post, and ball.
We evaluate the mean Average Precision (mAP) metric. In order to evaluate the effect of the different augmentation techniques, we use five different configurations of our dataset to train the previously described model:
\begin{enumerate}
  \item no augmentation at all (only the raw images generated by UERoboCup are used, this is the baseline)
  \item conventional augmentation without sun (all the augmentations described in \Secref{sec:augmentation} except for the sun patch augmentation)
  \item conventional augmentation (including the sun patch augmentation)
  \item only the image-to-image translation augmentation performed with MUNIT
  \item the image-to-image translation combined with all previously described conventional augmentations
\end{enumerate}

We train all five models with a subset of $8000$ images and $2000$ images as test set with the same augmentations in a batch size of $128$ using the Adam optimizer \cite{kingma_2014} with an initial learning rate of $0.1$.
We decay the learning rate by $0.5$, if the loss on the validation set does not decrease for $10$ epochs and we terminate training after $20$ epochs without improvement, which was usually reached in less than $100$ epochs.

\begin{figure}
  \centering
  \includegraphics[width=\textwidth]{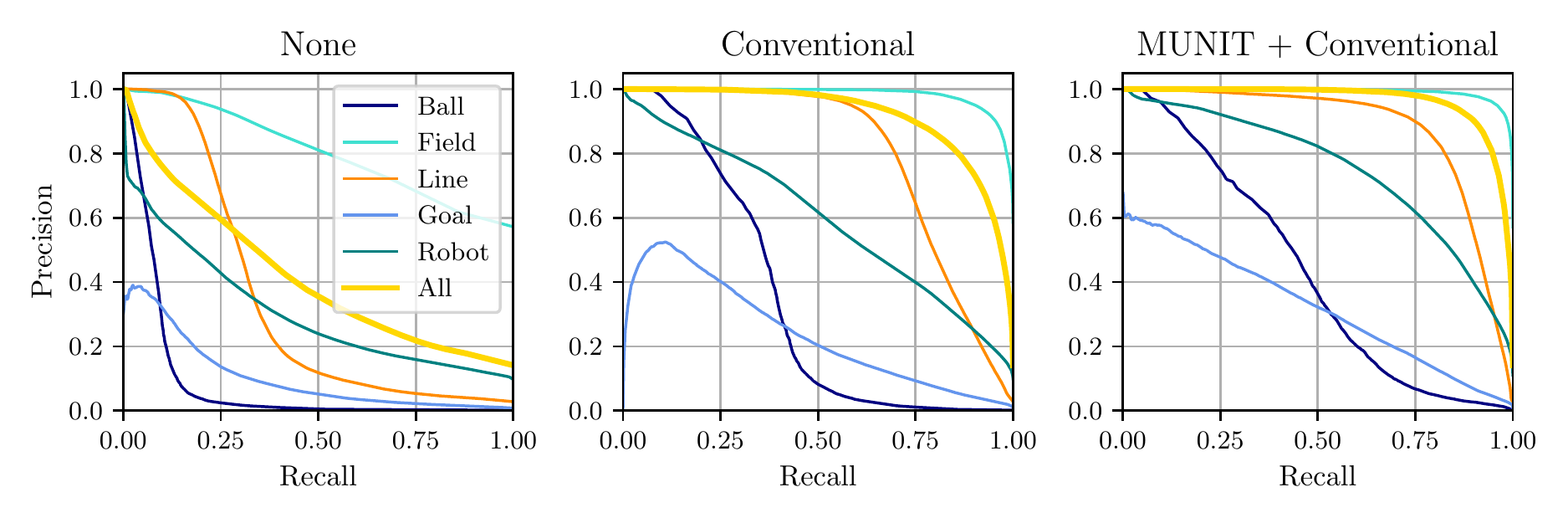}
  \caption{Precision-recall curves for a model trained without any augmentation (left), a model trained with conventional online augmentation as described before (middle), and a model trained with sim-to-real image translation and conventional augmentation (right).
A significant improvement can be seen due to the augmentation with MUNIT.}
  \label{fig:result}
\end{figure}

The different precision-recall curves are visualized in \Figref{fig:result}.
The mAPs for all models and all classes can be seen in \Tabref{tab:map}.
We report the precision-recall curves for the individual classes as well as micro-averaged for all classes.
Note that the computed mAP represents the classification performance of each individual pixel over all test images and not the classification performance of the object instances.

\begin{table}[t]
  \centering
  \caption{All mAP values for all tested models and all classes}
  \label{tab:map}
  \begin{tabular}{p{1.2cm}|>{\centering\arraybackslash}p{2cm}|>{\centering\arraybackslash}p{2cm}|>{\centering\arraybackslash}p{2cm}|>{\centering\arraybackslash}p{2cm}|>{\centering\arraybackslash}p{2cm}}
     & None      & Conventional without sun & Conventional & MUNIT & MUNIT and conventional \\
     \hline
    Ball & 0.0843 & 0.3927 & 0.3440 & \textbf{0.4577} & 0.4277 \\
Field & 0.8037 & 0.9846 & 0.9835 & 0.9877 & \textbf{0.9917} \\
Line & 0.3404 & 0.7866 & 0.7952 & 0.8745 & \textbf{0.8779} \\
Goal & 0.1024 & 0.1418 & 0.2367 & \textbf{0.3554} & 0.3207 \\
Robot & 0.3059 & 0.5361 & 0.5983 & 0.6529 & \textbf{0.7478} \\
All & 0.4203 & 0.9140 & 0.9165 & 0.9536 & \textbf{0.9647}

  \end{tabular}
\end{table}

\section{Discussion}
\label{s:discussion}
As expected, the model without any augmentation performs worst with an overall mAP of $0.4203$, which shows that reality gap is an issue for plain images generated in UERoboCup.
The ball and the goals are rarely detected with a slightly higher mAP for robots and lines.
Since a quantitative comparison with related works \cite{schnekenburger_2017,dijk_2018,szemenyei_2018} is not possible due to different metrics that do not capture the problem well, this model is used as baseline of what is possible with purely synthetic data generated in UERoboCup.

Just by performing basic augmentation (all augmentations mentioned above except for the sun patch augmentation), the performance of the model increases drastically to an overall mAP of $0.914$.
Most classes receive a significant increase in mAP, particularly the robot, line, and ball classes.
The goal classes' mAP only increase slightly.
Adding the sun patch augmentation increases the overall mAP again slightly to $0.9165$.
The sun patches augmentation helps improving the line, goal, and robot classification, but results in a drop in the ball class, while the field prediction stays about the same.
This demonstrates that the sun patch augmentation in fact helps.

When considering the model trained solely with data augmented with MUNIT, a clear rise in overall mAP to $0.9536$ can be seen, with an improvement for all classes.
Adding conventional augmentation to the MUNIT augmentation results in a slight overall mAP increase to $0.9647$.
While the ball and goal classes' mAPs drop again slightly, all other classes' performances increase.
This shows that our proposed method yields the desired result.
Particularly highly unbalanced classes benefit from the image-to-image translation augmentation.

The ball class is consistently the one with the lowest mAP, which is likely due to highly unbalanced training samples (with the ball being significantly smaller than any other objects).
The same applies to the goal post class.
The ball is one of the most challenging objects to detect in different lighting conditions since it is small, very close to the ground and throws shadows due to to its spherical shape.
Lastly, the model operates on a small resolution which makes it impossible to detect the ball at large distances.

\begin{figure}[t]
  \centering
  \begin{minipage}[t]{0.32\linewidth}
    \centering
    \includegraphics[width=\linewidth]{img/results/1_in.jpg}
  \end{minipage}
  \begin{minipage}[t]{0.32\linewidth}
    \centering
    \includegraphics[width=\linewidth]{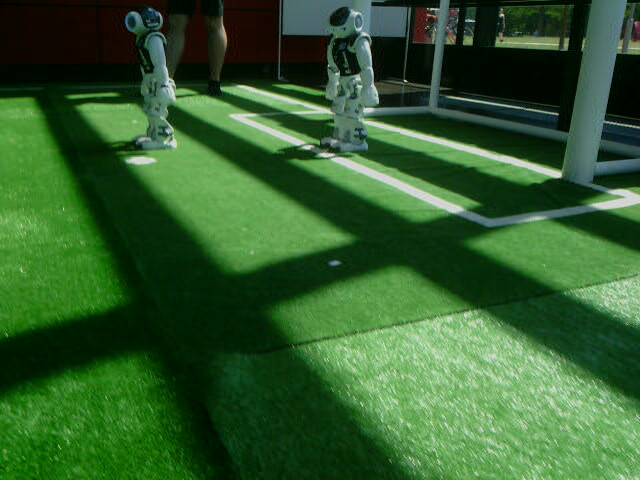}
  \end{minipage}
  \begin{minipage}[t]{0.32\linewidth}
    \centering
    \includegraphics[width=\linewidth]{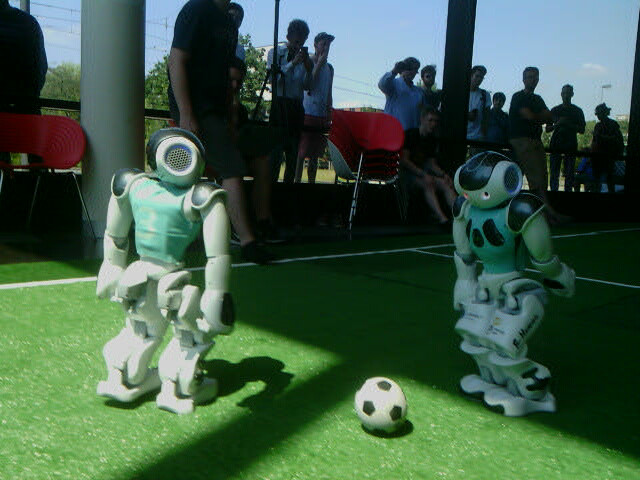}
  \end{minipage}
  \begin{minipage}[t]{0.32\linewidth}
    \centering
    \includegraphics[width=\linewidth]{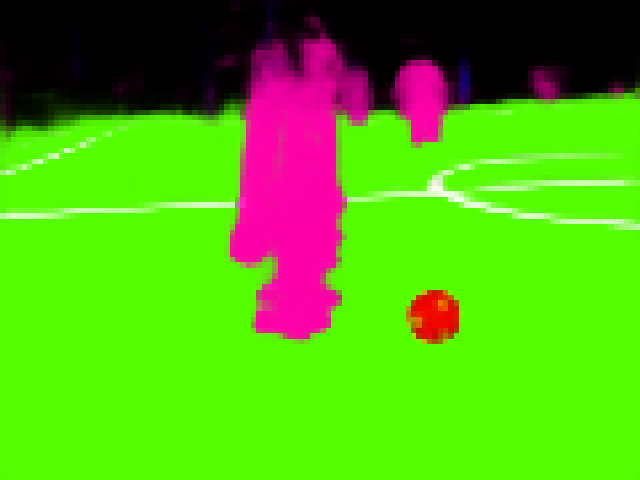}
  \end{minipage}
  \begin{minipage}[t]{0.32\linewidth}
    \centering
    \includegraphics[width=\linewidth]{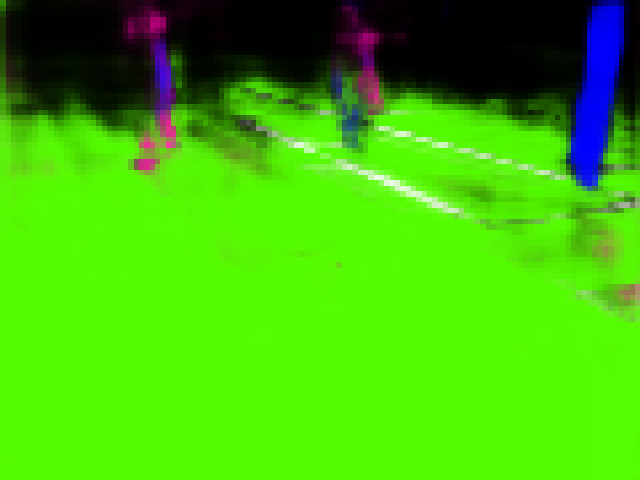}
  \end{minipage}
  \begin{minipage}[t]{0.32\linewidth}
    \centering
    \includegraphics[width=\linewidth]{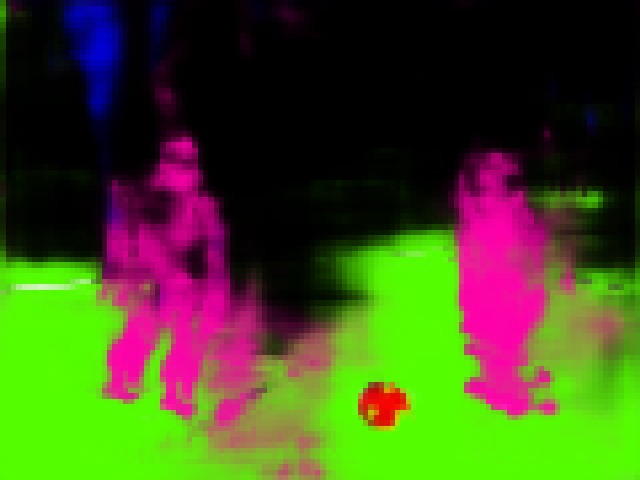}
  \end{minipage}
  \caption{Examples demonstrating the final performance of the neural network on an easy sample with constant lighting (left) and two hard samples with extreme lighting conditions (middle and right).
The best performing model was trained on the full dataset ($30000$ images) with a termination patiency of $40$ epochs.
The classes are encoded as follows: field (green), line (white), robot (pink), ball (red), goal post (blue), and background (black).
None of these images were used for the MUNIT training.
Despite the extreme lighting conditions, the segmentation performs reasonably well and even underrepresented classes such as the ball are mostly detected successfully.
}
  \label{fig:images_results}
\end{figure}

Despite its little size with only $12909$ trainable parameters (opposed to $300000$ parameters in \cite{schnekenburger_2017}), our proposed model trained with our synthetically generated training data seems to perform better than the approaches proposed by van Dijk \etal \cite{dijk_2018} and by Szemenyei \etal \cite{szemenyei_2018}.
A quantitative comparison is difficult due to the differing capabilities of the models.
While our model operates at a low resolution of only $120 \times 88$ pixels, van Dijk \etal operates at QVGA resolution and Szemenyei \etal use QQVGA resolution while Schnekenburger \etal \cite{schnekenburger_2017} use an image of $640 \times 512$ as input.
In contrast to van Dijk \etal, our model is successful in predicting multiple classes at once.
In contrast to Szemenyei \etal, our model seems to predict a subjectively more precise multi-class classification for each independent prediction, which Szemenyei \etal solve with an expensive label propagation.
Due to the same reasons, a runtime comparison is difficult, as van Dijk \etal and Schnekenburger \etal utilize GPUs for the inference.
With $14$ ms, our model is faster than the fastest model proposed by Szemenyei \etal ($22$ ms + $170$ ms label propagation).

The segmentation model was applied to images recorded in extreme lighting conditions.
Multiple samples can be seen in \Figref{fig:images_results}.

\section{Conclusion}

We proposed a method for segmenting an image in real-time at a reduced resolution into five different classes by utilizing a deep learning approach. We generated all training data synthetically and evaluated the results on a test set made of manually labeled real data.

We demonstrated our proposed method of generating a dataset with image-to-image translation with publicly available simulation tools.
With this dataset, we showed that a small semantic segmentation model is capable of running in real-time on low-end hardware with while producing results that outperform related work.
In contrast to end-to-end solutions such as \cite{bousmalis_2017,bewley_2018}, which integrate the image augmentation into the model, our approach allows to generate a versatile and high quality dataset, which we share with the community\footnote{\url{https://sibylle.informatik.uni-bremen.de/public/datasets/semantic\_segmentation}}, without the need to have access to high performance GPUs required to generate such datasets or the knowledge to design image-to-image translation models.





\bibliography{bibliography}

\end{document}